\documentclass[10pt,twocolumn,letterpaper]{article}

\usepackage[table]{xcolor}
\usepackage{algorithm}
\usepackage{algpseudocode}
\usepackage{array}
\usepackage[accsupp]{axessibility}  % Improves PDF readability for those with disabilities.

\usepackage{wacv}              % To produce the CAMERA-READY version
\newcommand{\modelname}{SurgTransfer\xspace}

\definecolor{wacvblue}{rgb}{0.21,0.49,0.74}
\usepackage[pagebackref,breaklinks,colorlinks,allcolors=wacvblue]{hyperref}
\usepackage[table]{xcolor}

\def\wacvPaperID{982} % *** Enter the WACV Paper ID here
\def\confName{WACV}
\def\confYear{2027}

\title{SurgTransfer: Multi-Step Latent Action Modeling for Action Transfer in Surgical World Model}

\author{
Saurabh Koju$^{1}$\thanks{These authors contributed equally to this work.} \quad
Love Panta$^{6}$\footnotemark[1] \quad
Deepak Thapa$^{1}$ \quad
Rudra P. K. Poudel$^{2}$ \quad
Prajwal Ghimire$^{3}$ \\
Angus Watson$^{6}$ \quad
Yash Raj Shrestha$^{5}$ \quad
Prashnna Kumar Gyawali$^{4}$ \quad
Binod Bhattarai$^{1,6}$
\vspace{0.5em} \\
$^{1}$NAAMII \quad
$^{2}$Toshiba Europe \quad
$^{3}$King's College London \\
$^{4}$West Virginia University \quad
$^{5}$University of Lausanne \quad
$^{6}$University of Aberdeen \\
{\tt\small saurabh.koju@naamii.org.np, \{l.panta.26, binod.bhattarai\}@abdn.ac.uk}}

\begin{document}
\maketitle

\begin{abstract}
Surgical world models have the potential to provide intra-operative guidance, accelerate risk-free training and facilitate transfer of expert surgical maneuvers. However, the use of world models for transferring surgical expertise has been largely unexplored. Moreover, recent approaches in this domain have mainly relied on ground truth kinematics actions or expert annotations for specific tasks, limiting their applicability beyond the narrow range of tasks with ground-truth annotations. In contrast, existing works in the natural domain have shown promising results in learning latent actions solely from unlabeled videos and transferring the learned actions across contexts. We observe that this paradigm is short-sighted for the surgical domain, failing to capture the hierarchical and goal-directed structure of surgical procedures. Thus, we propose a novel multi-step action learning framework to learn context-invariant latent actions that are easily transferable across various surgical contexts. We perform extensive experiments on the publicly available datasets, SurgToolLoc-2022 and PitVis-2023, verifying our approach with both quantitative metrics and qualitative evaluation by surgical experts. We have published our code and the implementation details at: \url{https://github.com/bhattarailab/Surgical-Vision-World-Model}.
\end{abstract}    
\section{Introduction}
\label{sec:intro}

\begin{figure}
    \centering
    \includegraphics[width=0.9\linewidth]{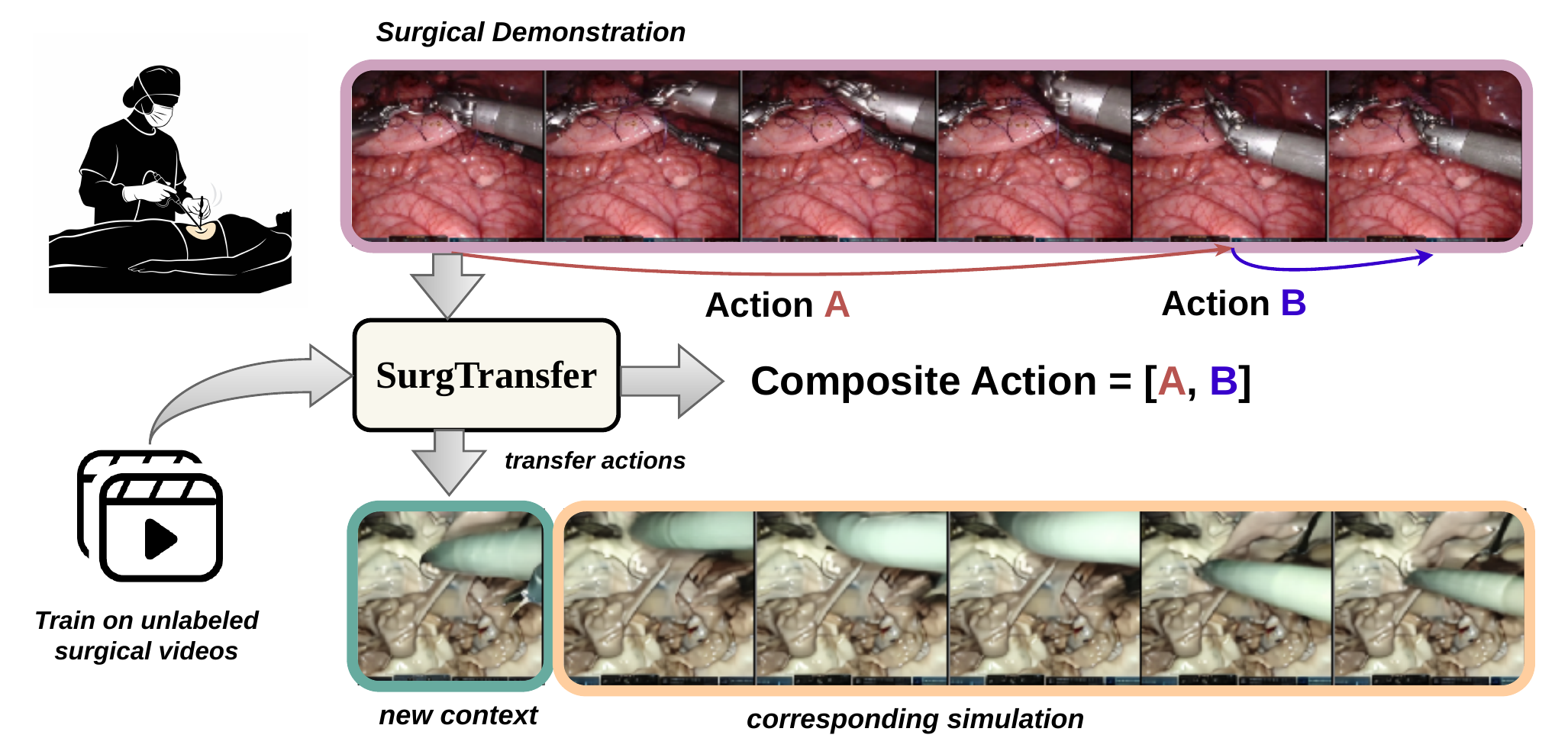}
    \caption{Our framework learns composable actions that can be transferred from demonstrations to novel surgical contexts.}
    \label{fig:introduction}
\end{figure}

World models learn an interactive simulation of an environment by predicting future observations conditioned on current states and actions~\cite{world_model_neurips_2018,hafner2020mastering,recore_cvpr_2024}. Such models hold significant promise for surgical applications, where they could enable intra-operative guidance, pre-operative planning, and risk-free simulation of procedural decisions.  For example, a surgeon can engage in risk-free interactions with the model to simulate the predicted outcomes of proposed surgical plans.  A trainee can explore alternative surgical strategies and rehearse decision-making without patient risk. Beyond simulation, learning action representations that are invariant to surgical contexts~\cite{koju2025surgical} would allow surgical skills demonstrated in one case to be transferred to anatomically different target cases, enabling skill reuse and training as shown in Figure~\ref{fig:introduction}. Despite this promise, transferable surgical world models have remained unexplored, and this paper makes the first contribution to our knowledge in learning transferable latent actions from unlabeled surgical videos.

Recent works that explore world models in surgical vision depend on explicit action supervision. Many methods rely on robotic actuator signals recorded from real or simulated systems~\cite{lin2025visuomotor,lin2024world,zbinden2025cosmos}, while others utilize textual descriptions as alternative representations of action~\cite{biagini2025hierasurg,turkcan2025towards}. Such annotations are scarce, noisy, and difficult to scale to general surgical procedures. To address action scarcity, several works in natural domains and  robotics~\cite{bruce2024genie,gao2025adaworld,zhang2024prelar} propose learning latent actions in an unsupervised manner. 
Surgical Vision World Model (SurgWM)~\cite{koju2025surgical} was the first work to extend this strategy to the domain of surgery. Our current work extends this work by studying \textit{transferring} latent actions across surgical contexts for which we find the existing paradigms myopic. Existing latent action models only infer actions between consecutive frames, which is not as well suited to surgical procedures, which are inherently hierarchical and goal-directed ~\cite{kim2025srt}. 

We address these limitations by introduce 
\emph{Multi-Step Latent Action Model (LAM)} (See Sec.~\ref{sec:Multi-step LAM}) designed to infer not only the action between consecutive frames but composite actions that span longer time frames. 
This design choice provides two key advantages: (a) It aligns naturally with the multi-frame prediction paradigm of video diffusion models, avoiding the progressive deterioration associated with autoregressive rollouts. 
While predicting multiple frames at a time is common for video diffusion models, world models generally predict single frame at a time to condition on actions between frames. However, since our transfer framework admits conditioning on multiple actions at a time, we can use the multi-frame generation capabilities of video diffusion models.
(b) Additionally, it encourages the model to encode coarser, composable latent actions. Our experiments demonstrate that, when transferred to a new target context, these high-level actions can better preserve their semantic consistency.

We summarize our key contributions below:
\begin{itemize}
    \item We introduce a general surgical procedure transfer framework: \modelname that is capable of transferring surgical actions across contexts.
    
    \item We propose a novel multi-step latent action model to learn composite latent actions from unlabeled surgical videos.
    
    \item We extensively ablate our method on different datasets and evaluate it both quantitatively, and qualitatively by surgical experts.
\end{itemize}

\section{Related Work}
\label{sec:literature}

\subsection{General World Models}
In recent years, there has been a significant leap in world modeling driven by large-scale internet videos, particularly for Embodied AI applications \cite{long2025survey}. According to \cite{li2025comprehensive}, world models can be categorized along three core dimensions: decision coupling, temporal reasoning, and spatial representation, which provide a fundamental framework to compare and contrast recent advancements.

Decision-coupled world models, such as \cite{schmidt2023learning, zhang2024prelar}, learn implicit action representations directly from videos without explicit control inputs. Building upon this line of work, Ye et al. \cite{ye2024latent} extend \cite{schmidt2023learning} by introducing visual-language-action models that integrate video observations and language instructions through behavior cloning. Similarly, \cite{ren2025videoworld} explores knowledge learning such as basic rules, reasoning and planning through compact representation of visual changes. In contrast, general-purpose world models \cite{gao2024vista, bruce2024genie, gao2025adaworld} are designed as task-agnostic simulators, emphasizing transferability and generalization over domain-specific tasks \cite{li2025comprehensive}. For instance, Genie \cite{bruce2024genie} represents the first interactive world model trained in a fully unsupervised manner. It discretizes latent actions into a set of codebooks to ensure human playability and maintain controllability. Closest to our work is AdaWorld \cite{gao2025adaworld}, which introduces an adaptable world model framework based on continuous latent actions, enabling transfer across variable contexts in natural and video game domain. We extend this work by learning actions at a coarser level that better suits the hierarchical nature of surgery.

\subsection{Surgical World Models}
World modeling has also begun to pave the way toward surgical dynamics modeling. For instance, \cite{lin2024world, lin2025visuomotor} introduced a DreamerV2-based \cite{hafner2020mastering} world model designed to learn visuomotor policies for surgical grasping tasks. Their approach leverages target object segmentation and the corresponding depth information to guide the robot gripper during manipulation. Notably, the predictions of the world model are conditioned on robotic actions available either in simulation or with real robots.

In contrast, works such as \cite{turkcan2025towards, biagini2025hierasurg} base their predictions on high-level descriptions of actions. Rapuri et. al.~\cite{saw} makes use of additional conditioning signals in the form of affordance and tool trajectory. Turkcan et. al. \cite{turkcan2025towards} developed suturing world models that use diffusion to generate videos of suturing tasks conditioned on sub-stitch action annotations. Similarly, Biagini et. al. \cite{biagini2025hierasurg} synthesize laparoscopic videos conditioned on hierarchical representation of the scene in the form of: surgical phase, action triplet and segmentation of the first frame. 

COSMOS-Surg-dVRK \cite{zbinden2025cosmos}, represents the first foundation world model based on Cosmos WFM \cite{agarwal2025cosmos} developed for evaluating the policy of the surgical robot. The model is obtained by finetuning Cosmos-Predict2-2B-Video2World on video-action paired samples derived from cholecystectomy dataset. While this model effectively learns surgical scene dynamics and action-conditioned predictions, it again makes use of ground truth robotic actions. 
Conversely, Cosmos-H-Surgical~\cite{cosmoshsurgical} attempts to circumvent ground-truth requirements by using an Inverse Dynamics Model (IDM) to generate pseudo-kinematics. However, this approach remains fundamentally limited as it still relies on kinematic action labels for training the IDM, restricting its efficacy to the specific robot embodiment used during data collection. SurgWM~\cite{koju2025surgical} learns implicit action representations from videos but does not investigate their transferability across surgical contexts.
\section{Methodology}
\label{sec:method}

\begin{figure*}[t]
    \centering
    % ---------------- FIGURE 2 ----------------
    \begin{minipage}[t]{0.60\linewidth}
        \centering
        \begin{subfigure}[b]{0.32\linewidth}
            \centering
            \includegraphics[width=\linewidth]{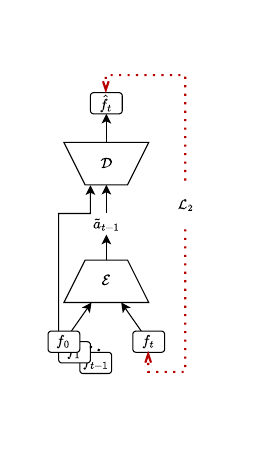}
            \vspace{-5mm}
            \caption{Autoregressive}
            \label{fig:autoregressive_lam}
        \end{subfigure}
        % \hfill
        \begin{subfigure}[b]{0.32\linewidth}
            \centering
            \includegraphics[width=\linewidth]{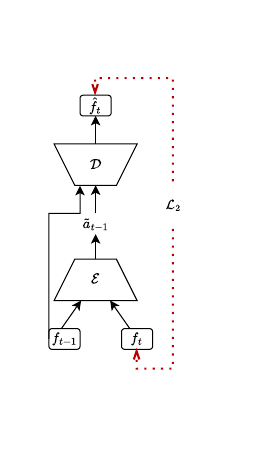}
            \vspace{-5mm}
            \caption{AdaWorld}
            \label{fig:adaworld_lam}
        \end{subfigure}
        % \hfill
        \begin{subfigure}[b]{0.32\linewidth}
            \centering
            \includegraphics[width=\linewidth]{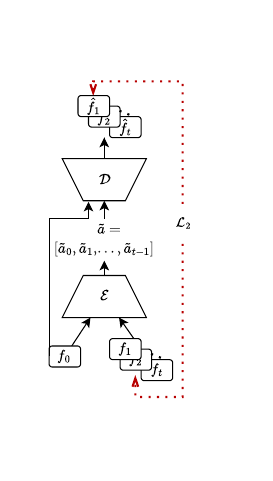}
            \vspace{-5mm}
            \caption{\modelname (Ours)}
            \label{fig:surgworld_lam}
        \end{subfigure}
        \caption{Comparison of different Latent Action Models. SurgTransfer (multi-step LAM) learns composable action i.e $\hat{a}$ that encodes long horizon changes.}
        \label{fig:lam_comparison}
    \end{minipage}
    \hfill
    % ---------------- FIGURE 3 ----------------
    \begin{minipage}[t]{0.365\linewidth}
        \centering
        \includegraphics[width=\linewidth]{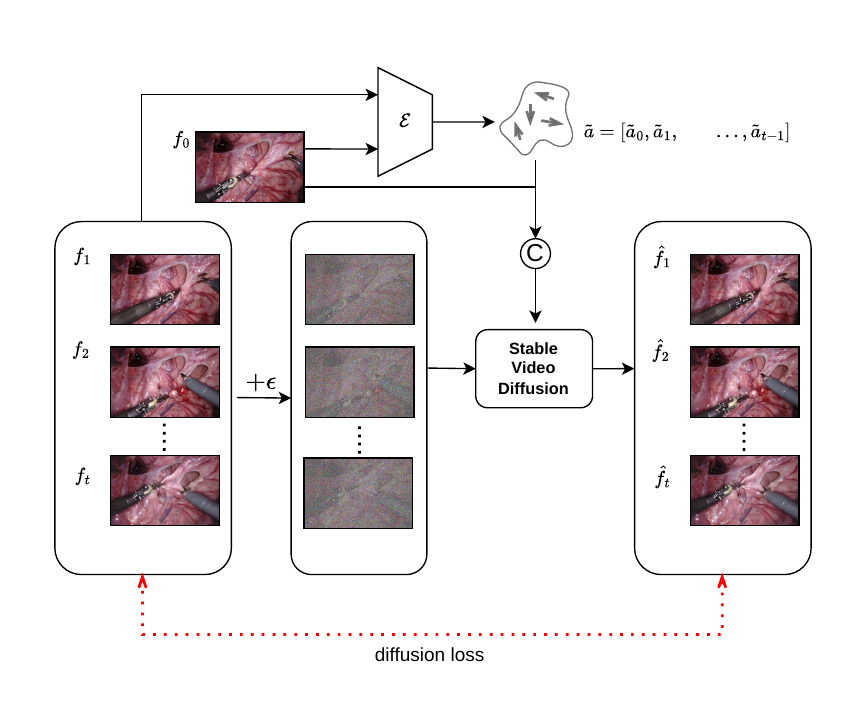}
        \caption{\modelname dynamics model conditions on the composite action representation.}
        \label{fig:diffusion}
    \end{minipage}
\end{figure*}

We study transferring the learned latent action $a_{s}$ from a given surgical source video $f_{s,0:T}$ to a given target surgical context, which is represented by a single frame $f_{c,0}$. This produces $f_{c,1:T}$, the predicted outcome of applying the source actions on the target context. To achieve this, following previous unsupervised works in world modeling \cite{bruce2024genie, gao2025adaworld}, we employ two major components: a latent action model (LAM) that learns latent actions between frames, and a dynamics model that predicts future frames conditioned on past observations and actions. However, in contrast to previous works, we learn a single composite latent action between $f_{s,0}$ and $f_{s, 1:T}$ during training, which can be thought of as a coarser action that captures the overall dynamics in the video. We will look at these components in detail in the subsequent sections.

\subsection{Latent Action Model (LAM)}
First, we will formulate the current latent action models in the literature under a common framework. Note that, even though a single observation typically refers to a single frame in the literature, for the purposes of this section, we will take the convention to use an observation to refer to either a single frame or a series of contiguous frames depending on the type of LAM used. 

A latent action model consists of an encoder $\mathcal{E}$ and a decoder $\mathcal{D}$. The encoder takes as inputs observations $o_{t-1}, o_{t}$ to infer the action between the observations as $\tilde{a}$ i.e. $\tilde{a} = \mathcal{E}\left(o_{t-1}, o_{t}\right)$.
The decoder takes observation $o_{t-1}$ and the inferred action $\tilde{a}$ to predict observation $\hat{o}_{t}$ i.e. $\hat{o}_{t} = \mathcal{D} \left(o_{t-1}, \tilde{a}\right)$.

The model is trained to minimize the reconstruction loss between $o_{t}$ and $\hat{o}_{t}$. The latent action $\tilde{a}$ can be either \textbf{discrete} \cite{bruce2024genie, ren2025videoworld}, in which case the model is trained with a VQ-VAE objective \cite{van2017neural}, or \textbf{continuous} \cite{gao2025adaworld}, in which case the model is trained with a VAE objective \cite{vae}. Previous works \cite{bruce2024genie} typically use a number of historical frames $f_{t-k:t-1}$ as $o_{t-1}$ and a single future frame $f_{t}$ as $o_{t}$. We refer to models using $k = t$ as Autoregressive LAM models, and $k = 1$  as AdaWorld \cite{gao2025adaworld} LAM model as shown in figures \ref{fig:autoregressive_lam} and \ref{fig:adaworld_lam}. For continuous actions, the corresponding objective function can be defined as:

\begin{equation}
    \begin{split}
    \mathcal{L}_{\mathrm{LAM}} =
    &\Big\| 
        f_{t} -
        \mathcal{D}\big(
            f_{t-k:t-1},\,
            \underbrace{\mathcal{E}\big(f_{t-k:t-1}, f_{t}\big)}_{\tilde{a}\,=\,a_{t-1}}
        \big)
    \Big\|^2 \\
    &+ \beta\, D_{\mathrm{KL}}\big(q\left(\tilde{a}\right),\, \mathcal{N}\left(0, \mathcal{I}\right)\big) 
    \end{split}
    \label{eq:lam_old}
\end{equation}

where, $t = 1, 2, ..., T$ 

However, since the model only predicts one frame $f_{t}$ at a time, the model only learns to encode the fine-grained motion between frames $f_{t-1}$ and $f_{t}$ as action. Moreover, because of the temporal redundancy in the video, the two frames have a high degree of overlap, making the prediction task easier. Thus, the learned actions are inherently myopic, failing to capture the long-range dynamics in surgical videos.

\subsubsection{Multi-step LAM}
\label{sec:Multi-step LAM}
The discussion above motivates us to reframe the action learning framework as predicting $n$ frames at a time conditioned on a single frame of context, so that it models long range interactions in the environment. This forces the model to learn higher level actions rather than low level motion encodings. Thus, we can introduce our latent action model by setting $o_{t-1} = f_0$, and $o_{t} = f_{1:t}$ as shown in Figure \ref{fig:surgworld_lam}. The corresponding objective then becomes:

\begin{equation}
    \begin{split}
    \mathcal{L}_{\mathrm{LAM}} =
    &\Big\| 
        f_{1:t} -
        \mathcal{D}\big(
            f_0,\,
            \underbrace{\mathcal{E}\big(f_0, f_{1:t}\big)}_{\tilde{a}\, = \,[a_0, a_1, ..., a_{t-1}]}
        \big)
    \Big\|^2 \\
    &+ \beta\, D_{\mathrm{KL}}\big(q\left(\tilde{a}\right),\, \mathcal{N}\left(0, \mathcal{I}\right)\big)
    \end{split}
    \label{eq:lam}
\end{equation}

We minimize this objective for all $t = 1, 2, .., T$ using causal architecture of the encoder and decoder. For example, to reconstruct $f_j$, the decoder must use the initial context frame $f_0$ and the corresponding actions $a_{0:j-1}$, while to reconstruct $f_{j+1}$, the decoder must use the actions $a_{0:j}$. Here, the additional action $a_j$ should still encode the change between the frames $f_j$ and $f_{j+1}$ like the previous LAM models. However, since the decoder only has access to the first frame the model should learn the actions $a_{0:j-1}$ (the change between $f_0$ and $f_j$) and $a_j$ such that they can be composed together to encode the overall change from $f_0$ to $f_{j+1}$. Our experiments show that the obtained actions show higher invariance to a change of context.

Here, we have used continuous actions to represent more nuanced actions possible in surgery. We also employ the $\beta$-term introduced in $\beta$-VAE \cite{higgins2017beta}, to balance the expressiveness and the information content of the action.

\subsection{Multi-Step Dynamics Model}
A dynamics model predicts observation $o_{t}$ given observation $o_{t-1}$ and action $\tilde{a}$. Similar to LAM, we can define different dynamic models in terms of the past observations they condition on and the future predictions they make. 

Existing dynamics models \cite{bruce2024genie, gao2025adaworld} are autoregressive, where $o_{t}$ represents $f_{t}$, and $o_{t-1}$ represents $f_{0:t-1}$. Such models, however, are prone to \textit{autoregressive deterioration}, as they condition on their own generated frames during inference as shown in Figure \ref{fig:same_transfer}. Recognizing that our action transfer framework only requires the dynamics model to predict a full video from a single image rather than modeling every inter-frame transition, we instead formulate our task as predicting video given an image and a composite action, i.e. with $o_{t-1} = f_0$ and $o_{t} = f_{1:t}$ as shown in Figure \ref{fig:diffusion}.

To this end, we finetune the publicly available image-to-video Stable Video Diffusion (SVD) \cite{blattmann2023stable} model, which is shown to be action-controllable, multi-view consistent, and generalizable \cite{voleti2024sv3d}. The UNet \cite{unet} is trained to also condition on the action $\tilde{a}$ in addition to the CLIP \cite{radford2021learning} image embedding of the conditioning frame and the channel-wise concatenation of the noise-augmented conditioning frame used by the original SVD. We adopt the EDM framework \cite{karras2022elucidating} for finetuning, following the original SVD training setup. A simplified form of the training objective of the diffusion dynamics model can be expressed as:
\begin{equation}
    \mathcal{L}_{\mathrm{dynamics}} = \mathbb{E}_{v, \epsilon, t}\left[\big\|(v-\hat{v}\ (v_{noised},f_0, \tilde{a}, t))\big\|^2\right]
\end{equation}
where $v$ denotes the frames $f_{1:t}$, and $\hat{v}$ denotes its denoised prediction by the model. 

\section{Experiments}
\label{sec:experiments}
% \subsection{Implementation Details}

\subsection{Datasets and Baselines}
We evaluated \modelname on two of the publicly available surgical datasets, SurgToolLoc-2022 \cite{zia2023surgical} and PitVis \cite{das2025pitvis}. The SurgToolLoc\allowbreak-2022 comprises 24,695 video clips of robot-assisted surgical training exercises performed with the da Vinci Surgical System. 
Whereas, PitVis contains 25 videos with an average length of over an hour of endoscopic pituitary surgeries. 

AdaWorld \cite{gao2025adaworld} is currently the only framework that investigates transferability in world modeling, and thus serves as our primary baseline for evaluating video generation. Other works are not directly comparable for action transfer evaluation. Additionally,
we have reported the relevant metrics from SurgWM \cite{koju2025surgical} which conducted their experiments in the same dataset. We also evaluate different combinations of latent action and dynamics models. For LAM, we study three variants: Autoregressive, Adaworld, and \modelname as shown in Figure  \ref{fig:lam_comparison}. Autoregressive LAM can either be discrete \cite{bruce2024genie} or continuous. 
Following ~\cite{gao2025adaworld}, we also employ the UniMatch ~\cite{xu2023unifying} model to use the optical flow between frames as actions for an additional baseline. For dynamics models, we compare the AdaWorld's autoregressive and \modelname's multi-step diffusion models. We train all of these variants under the same protocol as \modelname and assess their performance on the SurgToolLoc-2022 dataset in terms of cross-context adaptation. We only compare AdaWorld and \modelname on the PitVis dataset owing to computational constraints. We use 80:20 train:test split on both datasets.

\subsection{Implementation Details} 
We utilize the ST-Transformer \cite{xu2020spatial} for both encoder and decoder of the latent action model. Both modules comprise of 12 blocks of spatial and temporal attention layers with 512 channels and 8 attention heads. We use 32 dimensional latent actions with $\beta$ set to $ 2 e-4$.
% \textcolor{red}{ata? } For the segmentation objective, we set the weight $\alpha$ to $0.05$.
The latent action models are trained for 100k steps on 2 NVIDIA A100 GPUs. Both LAMs and dynamics models are trained with window size 8, unless otherwise mentioned. The diffusion model is finetuned for 80k steps on 8 NVIDIA L40 GPUs. For generating frames beyond the window size that was used to train the dynamics model, the last generated frame is used as the new reference frame. We perform 25 sampling steps for all the diffusion models. All metrics are reported at prediction horizon of length 14.

\subsection{Evaluation measures}
We provide both quantitative and qualitative evaluation protocols to verify the 
effectiveness of the proposed approach.
Our quantitative evaluation focuses on whether the models can learn \textit{transferable} latent actions. To achieve this, we evaluate the models along two dimensions: the visual plausibility of videos generated through action transfer and the semantic consistency of the latent actions. For the former, we calculate the Content-Debiased Fréchet Video Distance (CD-FVD) \cite{ge2024content} between the source and transferred videos. 
Compared to Fréchet Video Distance (FVD) \cite{unterthiner2018towards}, CD-FVD places greater emphasis on motion dynamics rather than static visual content. To perform better on this metric, the model should maintain high visual fidelity while faithfully transferring the dynamic content of the source videos. We also use Embedding Cosine Similarity (ECS) to complement CD-FVD with frame-level measurements following AdaWorld. To further ensure that our quantitative evaluation measures not only  visual plausibility but also semantic consistency during action transfer, we augment these metrics with Action Cycle Consistency metric introduced by~\cite{inthewild}. It re-infers the latent actions from the transferred videos and then applies them to the source context, which should closely reproduce the source video if the action transfer works well. Following~\cite{inthewild}, we report the LPIPS between the original source video and the cyclically reproduced source video.

For PitVis, where step annotations (e.g. nasal corridor creation, tumor excision, debris clearance, etc.) are available, source and target segments are sampled from different surgical videos in such a way that they represent the same surgical step. For SurgToolLoc where step annotations are not available, each surgical clip, which is 30 seconds long, is divided into two temporal segments, forming source–target pairs. Latent actions extracted from the source segments were transferred to the target context image to assess whether the model maintains consistent and context-aware behaviors across similar surgical scenes. This pairing strategy enables a fair comparison by preserving procedural similarity while varying contextual dynamics such as instrument orientation and tissue motion. 

We also measure reconstruction fidelity which measures the model’s ability to
reproduce video sequences conditioned on implicit latent actions inferred by the LAM. For this purpose we adopt frame-level metrics such as PSNR \cite{hore2010image}, SSIM \cite{wang2004image}, LPIPS \cite{zhang2018unreasonable}, and CMMD \cite{jayasumana2024rethinking}. Unlike FID \cite{heusel2017gans}, which depends heavily on the Inception backbone and sample size, CMMD offers a more robust alternative for evaluating video generation quality. It leverages CLIP embeddings in conjunction with MMD distance to compute the score. In our case, we employ PeskaVLP \cite{yuan2024procedure} as the embedding model, which is pretrained on a large-scale surgical video–language dataset. This choice also helps mitigate potential distribution mismatches within the surgical domain.

\subsection{Clinical Evaluation}
Following \cite{chen2025far}, we also employ a clinically-grounded framework that use expert evaluations on different criteria that are specifically designed for evaluating world models on surgical plausibility necessary for meaningful clinical applications. This framework not only measures the visual plausibility of generations but also evaluates them on plausibility of instrument operation, environment feedback and surgical feedback. In addition to these four levels, we incorporate an additional category relevant to our action transfer framework: Action transfer plausibility. This is designed to evaluate the faithfulness in preserving the function and surgical intent of the transferred action. We assess a total of forty generated videos independently with five certified surgeons using a rigorous questionnaire.

\begin{table}[!htb]
\centering
\small
\caption{Clinical evaluation of \modelname and AdaWorld on different levels of surgical plausibility.}
\label{tab:clinical}
\setlength{\tabcolsep}{6pt}
\begin{tabular}{@{}l|c|c@{}}
\toprule
\textbf{Plausibility levels}
& \textbf{AdaWorld~\cite{gao2025adaworld}}
& \textbf{\modelname} \\
\midrule
Visual Perceptual$\uparrow$
& $2.65 \pm 0.30$ & $\mathbf{3.42 \pm 0.35}$ \\
Instrument Operation$\uparrow$
& $2.63 \pm 0.32$ & $\mathbf{3.27 \pm 0.37}$ \\
Environment Feedback$\uparrow$
& $2.64 \pm 0.31$ & $\mathbf{3.31 \pm 0.34}$ \\
Surgical Intent$\uparrow$
& $2.62 \pm 0.33$ & $\mathbf{3.23 \pm 0.35}$ \\
Action Transfer$\uparrow$
& $2.63 \pm 0.31$ & $\mathbf{3.15 \pm 0.38}$ \\
\bottomrule
\end{tabular}
\end{table}

As shown in Table~\ref{tab:clinical}, our model scores higher on all levels of plausibility than AdaWorld. Visual perceptual plausibility is high for both models, with a decreasing trend seen on higher levels of surgical plausibility, as observed by~\cite{chen2025far} as well. The most dramatic improvement is seen in visual perceptual plausibility reporting an increase of 29\% over the baseline. 
We measure inter-rater reliability using a two-way random-effects, absolute-agreement ICC for \(k\) raters. Averaged across the five plausibility levels, ICC(2,\(k\))~\cite{mcgraw1996forming} was 0.51 for AdaWorld and 0.54 for SurgTransfer.

\begin{figure*}[t]
    \centering

    %--- Left: first figure (2/3 width) ---
    \begin{minipage}[c]{0.60\linewidth}  % center vertically
        \centering
        \begin{subfigure}[b]{0.455\linewidth}
            \centering
            \includegraphics[width=\linewidth]{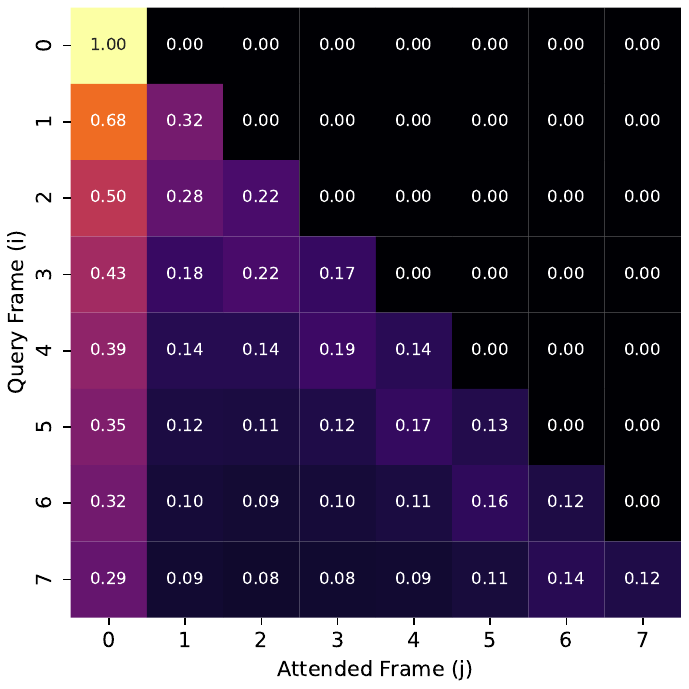}
            \caption{\modelname LAM model}
            \label{fig:surgworld_temp_att}
        \end{subfigure}
        \hfill
        \begin{subfigure}[b]{0.535\linewidth}
            \centering
            \includegraphics[width=\linewidth]{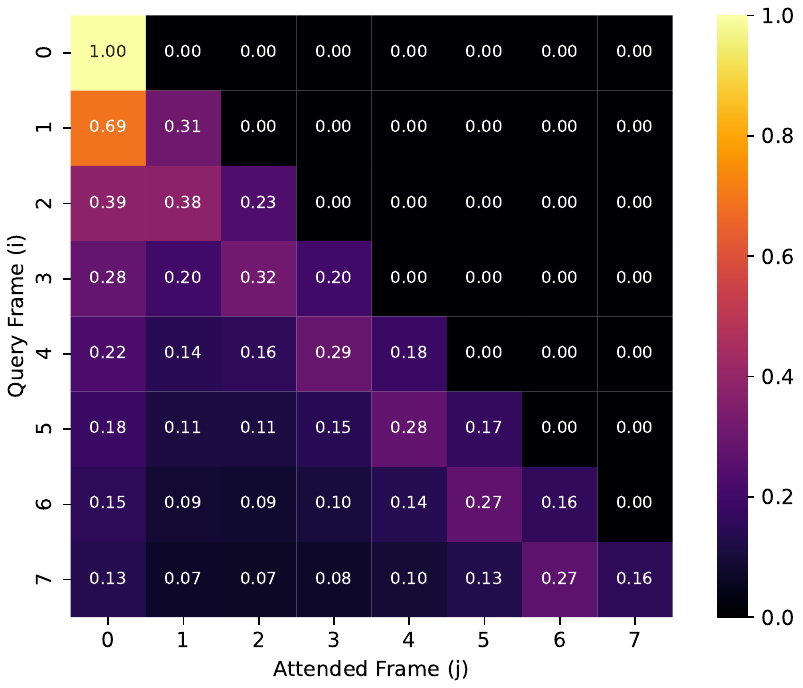}
            \caption{Autoregressive LAM model}
            \label{fig:genie_temp_att}
        \end{subfigure}
        \caption{Comparison of temporal attention maps between \modelname and Autoregressive LAM models. Autoregressive LAM shows greater reliance on the preceding frame. }
        \label{fig:temporal_comparison}
    \end{minipage}
    \hfill
    %--- Right: second figure (1/3 width) ---
    \begin{minipage}[c]{0.33\linewidth}  % center vertically
        \centering
        \includegraphics[width=\linewidth]{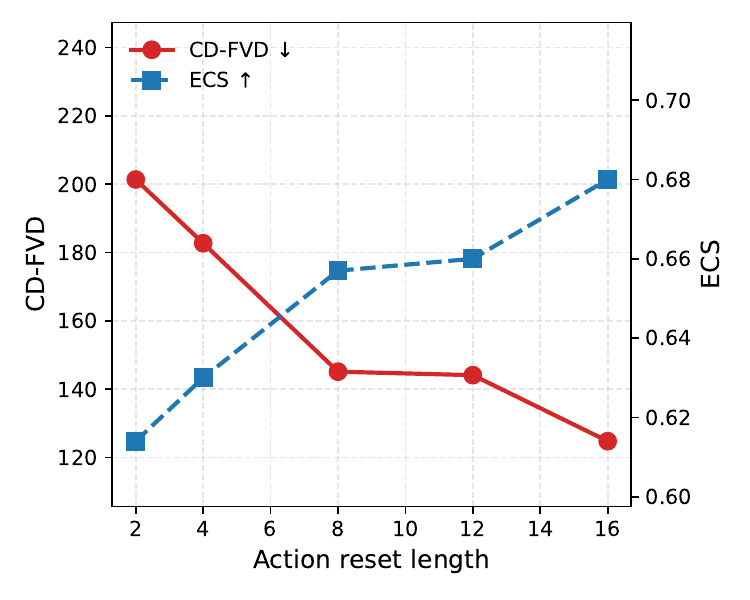}
        \caption{Increasing context length leads to improved performace in action modeling.}
        \label{fig:action_reset}
    \end{minipage}

\end{figure*}

\subsection{Qualitative Evaluation}
We visualize the action transfer capability of our model in Figure \ref{fig:same_transfer}. In each case, we use actions inferred from the source video by LAM and the starting context frame to roll out 14 more frames into the future. 
We visually inspect motion cues such as occurrences of temporal inconsistency between generated video and original video. 
We find that \modelname shows better scene consistency and generation quality while transferring actions than AdaWorld. Particularly, \modelname does not suffer as much from progressive deterioration that is seen in AdaWorld and also demonstrates better long-term temporal consistency by maintaining object identity when objects leave and re-enter the frame as in the third example.

\subsubsection{Latent Action Information Visualization}
\begin{figure*}[h!]
    \centering
    \begin{subfigure}[b]{0.48\linewidth}
        \centering
        \includegraphics[width=\linewidth]{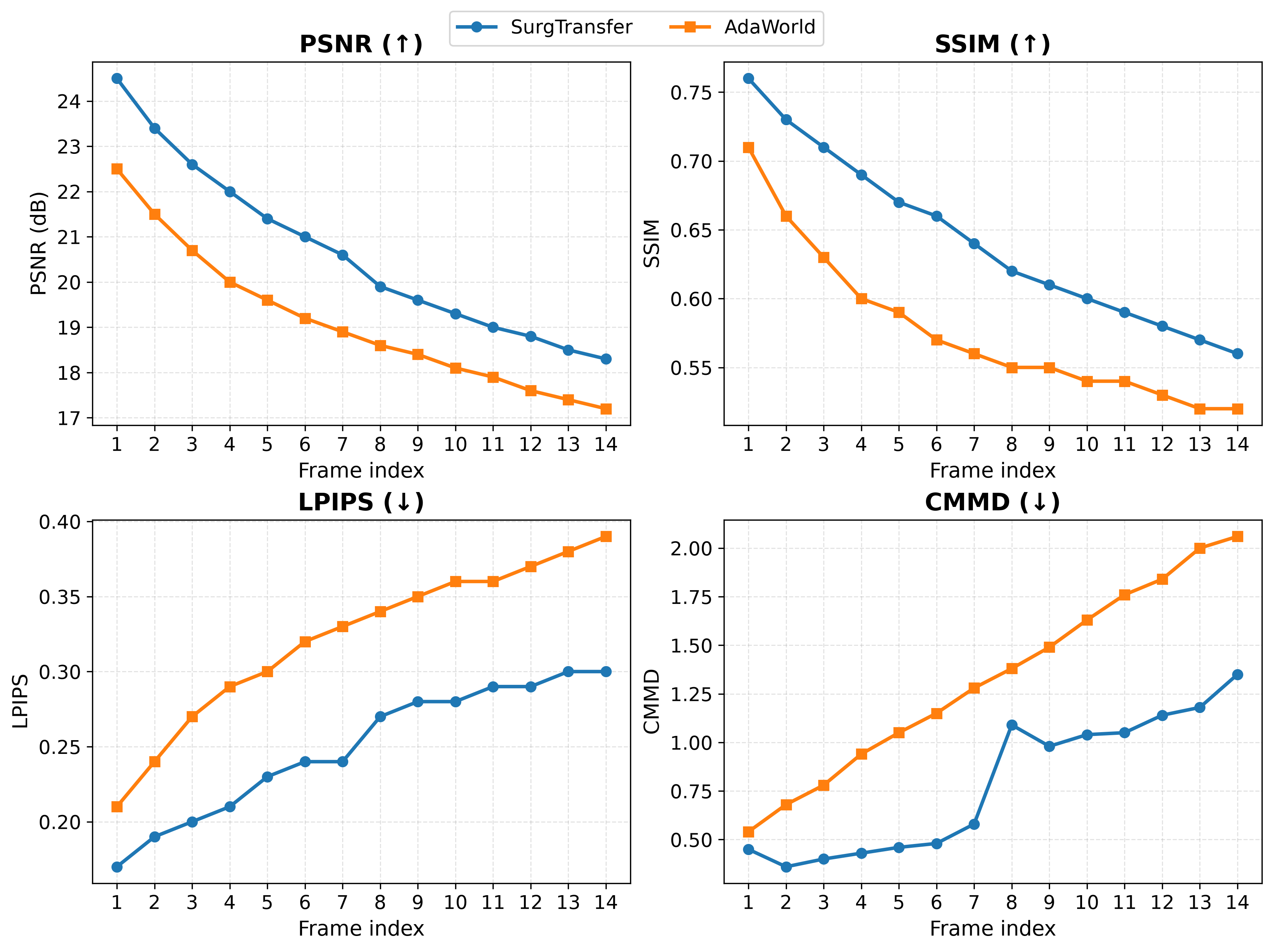}
        \caption{Comparison of visual per-frame metrics.}
        \label{fig:per_frame}
    \end{subfigure}
    \hfill
    \begin{subfigure}[b]{0.51\linewidth}
        \centering
        \includegraphics[width=\linewidth]{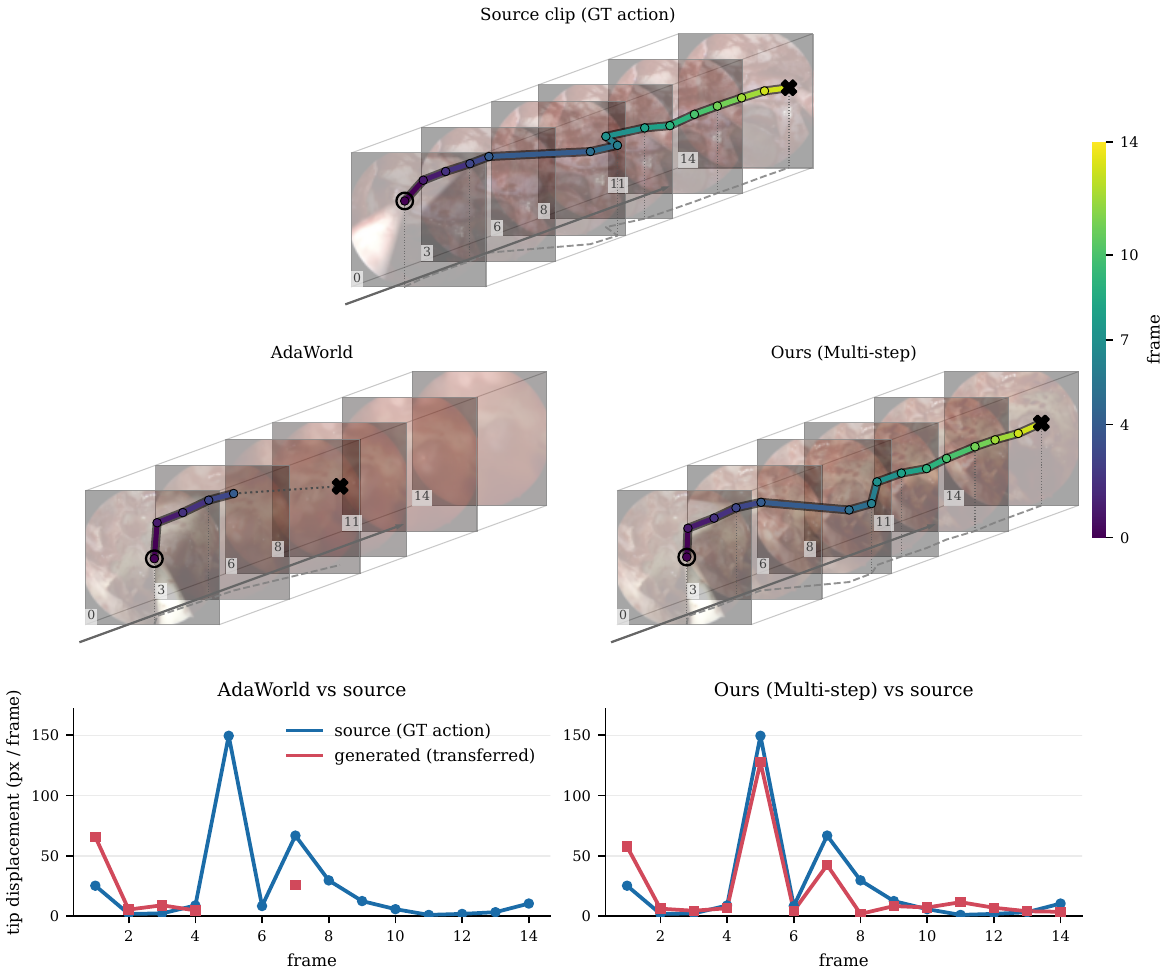}
        \caption{Visualization of instrument trajectories during action transfer.}
        \label{fig:latent_best}
    \end{subfigure}

    \caption{Comparison of SurgTransfer and AdaWorld for long-horizon prediction.}
    \label{fig:comparison}
\end{figure*}

Continuous latent actions offer limited interpretability 
regarding
the information 
they encode.
However, we can track and visualize the trajectory of the tip of the visible instruments in the source and transferred videos, as shown in Figure~\ref{fig:latent_best}, to gain partial insights into the information encoded by the latent actions.
% ,~\ref{fig:latent_good}, and ~\ref{fig:latent_more}. 
We follow the same protocol as above on the PitVis dataset to roll out 14 frames in the target context. 
The instrument tips were labeled manually. As can be seen from the figure, our LAM effectively captures the movements of the instrument.
% as shown in the figure. 
Adaworld, on the other hand, shows the instrument disappearing in frame 5 when a large displacement occurs. 
Further qualitative evaluation and analysis of the results are provided in the appendix.

\newcolumntype{L}[1]{>{\raggedright\arraybackslash}p{#1}}
\newcolumntype{C}[1]{>{\centering\arraybackslash}p{#1}}

\begin{table}[!htb]
\centering
\footnotesize
\caption{Comparison of \modelname against different baselines on Action Transfer. AR: autoregressive models. All LAMs use continuous actions unless otherwise specified.}
\label{tab:transfer_metrics}
\setlength{\tabcolsep}{2pt}
\renewcommand{\arraystretch}{1.1}
\begin{tabular}{@{}L{1.6cm}L{1.25cm}L{1.25cm}C{0.95cm}C{0.75cm}C{1.2cm}@{}}
\toprule
\textbf{Method} & \textbf{LAM} & \textbf{Dynamics}
& \textbf{CD-FVD}$\downarrow$ & \textbf{ECS}$\uparrow$
& \textbf{Cyclic Consistency}$\downarrow$ \\
\midrule

%--- SurgToolLoc ---
\multicolumn{6}{@{}c}{\textit{SurgToolLoc}} \\
AdaWorld~\cite{gao2025adaworld} & AdaWorld & AR & 341.76 & 0.53 & 0.321 \\
\cmidrule(l){2-6}
\textbf{\modelname} & Multi-step & Multi-step & \textbf{141.41} & \textbf{0.67} & \textbf{0.297} \\
\cmidrule(l){2-6}
Variants & Multi-step & AR & 338.12 & 0.53 & 0.367 \\
\cmidrule(l){2-6}
 & AR & Multi-step & 171.41 & 0.61 & 0.322 \\
\cmidrule(l){2-6}
 & Discrete, AR & Multi-step & 361.07 & 0.48 & 0.327 \\
\cmidrule(l){2-6}
 & Optical Flow & Multi-step & 155.92 & 0.60 & 0.321 \\
\midrule

%--- PitVis ---
\multicolumn{6}{@{}c}{\textit{PitVis}} \\
AdaWorld~\cite{gao2025adaworld} & AdaWorld & AR & 513.27 & 0.51 & 0.410 \\
\cmidrule(l){2-6}
\textbf{\modelname} & Multi-step & Multi-step & \textbf{157.34} & \textbf{0.66} & \textbf{0.301} \\
\bottomrule
\end{tabular}
\end{table}

\subsection{Quantitative Evaluations}
\noindent\textbf{Cross-context adaptation:}
Table ~\ref{tab:transfer_metrics} shows the results of cross context adaptation for different variants of world models. Here, frame-level metrics were not deemed as useful for evaluating cross-context adaptation due to content variation in source and target pair. 
We obtained significant improvements of our LAM models: lowering CD-FVD by 58\% in SurgToolLoc and by 69\% in PitVis,  and lowering Cyclic LPIPS by 7.48\% in SurgToolLoc and 26.59\% in PitVis.
We can also see the generation quality of our model in comparison to AdaWorld in Figure ~\ref{fig:same_transfer}. 

\noindent \textbf{Reconstruction Fidelity:} 
Additionally, we evaluate the models on reconstruction of the same video using the inferred ground truth actions from the LAM, to evaluate the information content captured by the latent actions and the resulting visual fidelity. 
We calculate several metrics to measure video generation quality of the different models in Table \ref{tab:reconstruction_metrics}. Since SurgWM also studied video fidelity for reconstruction task from inferred actions, we include the relevant metrics available from the paper.
We see a significant gap in performance between autoregressive and multi-step dynamics model for high quality fidelity generation. We also find that our LAM model improves the prediction ability of both dynamics models. We also see strong performance of the optical flow baseline on the reconstruction task. This can be attributed to optical flow being able to encode a large amount of information about the future frame in the action; a strategy that does not work well for cross-context action transfer.

\begin{table*}[t!]
\centering
\small
\caption{Comparison of \modelname against baselines on reconstruction for SurgToolLoc \cite{zia2023surgical}.}
\label{tab:reconstruction_metrics}
\setlength{\tabcolsep}{6pt}
\begin{tabular}{@{}l|c|c|c|c|c|c|c|c@{}}
\toprule
\textbf{Method} & \textbf{LAM} & \textbf{Dynamics} & \textbf{PSNR}$\uparrow$ & \textbf{SSIM}$\uparrow$ & \textbf{LPIPS}$\downarrow$ & \textbf{CMMD}$\downarrow$ & \textbf{CD-FVD}$\downarrow$ & \textbf{ECS}$\uparrow$ \\
\midrule
SurgWM~\cite{koju2025surgical}    & Discrete, AR & AR MaskGIT & 16.49               & 0.39                & -                   & -                   & -                   & -             \\
AdaWorld~\cite{gao2025adaworld}   & AdaWorld     & AR         & 19.098              & 0.576               & 0.321               & 1.23                & 384.205             & 0.576         \\
\textbf{\modelname}               & Multi-step   & Multi-step & \textbf{20.572}     & \textbf{0.635}      & \textbf{0.247}      & 0.713               & 176.133             & \textbf{0.747}\\
\cmidrule(l){1-9}
Variants                          & Multi-step   & AR         & 19.525              & 0.596               & 0.308               & 1.032               & 376.486             & 0.576         \\
                                  & AR           & Multi-step & 19.398              & 0.602               & 0.273               & 0.68                & 224.156             & 0.657         \\
                                  & Discrete, AR & Multi-step & 18.031              & 0.565               & 0.303               & \textbf{0.601}      & 374.51              & 0.482         \\
                                  & Optical Flow & Multi-step & 19.756              & 0.626               & 0.256               & 0.666               & \textbf{166.973}    & 0.72          \\
\bottomrule
\end{tabular}
\end{table*}

\subsection{Key-findings}
\begin{figure*}
    \centering
    \includegraphics[width=0.89\linewidth]{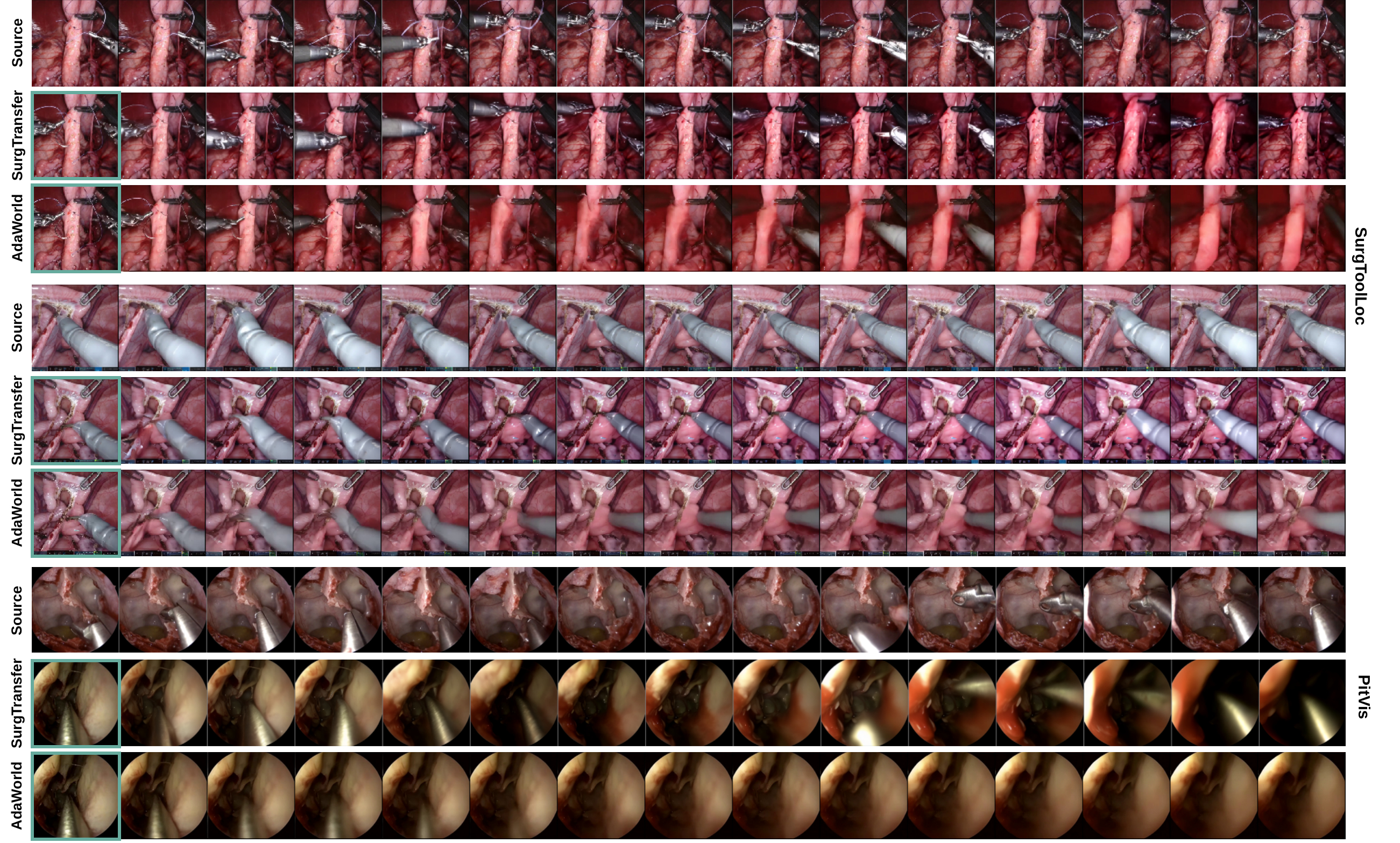}
    \caption{Action transfer: source represents the source clip from which the action is inferred, the first frame (highlighted in light green box) represents the context, rest of the frames are generated using the context and transferred actions}
    \label{fig:same_transfer}
\end{figure*}

\noindent \textbf{Key-finding 1: Multi-step dynamics model enables higher-fidelity and more consistent video generation.}
We can see improved performance of multi-step dynamics model in terms of both action transfer and reconstruction fidelity as shown in Table \ref{tab:transfer_metrics} and \ref{tab:reconstruction_metrics}. There exists significant gap from autoregressive dynamics model to multi-step dynamics model for both tasks. This is due to the autoregressive deterioration inherent in these models that results in lower quality generations as the prediction horizon increases.
Specifically, visual comparisons between AdaWorld and \modelname show that instrument tips frequently vanish, and tool motion exhibits significant deviations from the ground truth as can be seen in Figure \ref{fig:same_transfer}. \modelname also poses strong consistency both in spatial and temporal domain, highly matching the procedural dynamics between source and target surpassing AdaWorld in all examples.

\noindent \textbf{Key-finding 2: Better alignment of LAM model with diffusion model improves performance.}
We investigated the primary cause of the performance drop observed in autoregressive LAM compared to \modelname for the same multi-step dynamics model. 
Specifically, we analyzed how the current frame is influenced by its predecessor frames and how the LAM decoder enforces learning of desirable temporal changes. This was achieved by plotting the temporal attention maps for each frame, as shown in Figure~\ref{fig:temporal_comparison}. Our analysis revealed that autoregressive LAM relies more heavily on the immediate previous frame than on the initial frame, which likely leads to misalignment with the Multi-step dynamics model. In contrast, \modelname places greater emphasis on the initial frame, enabling better latent action encoding that ultimately enhances the diffusion model’s performance.

\noindent \textbf{Key-finding 3: \modelname LAM learns better temporal abstractions for action transfer.}
As can be seen from the Table \ref{tab:transfer_metrics}, the \modelname LAM captures actions that give higher fidelity generations and show higher context invariance regardless of the dynamics models used. This shows that our latent action model is better in capturing long horizon dynamics that is inherent in the surgical procedures. LAMs that predict single frame into the future i.e. Autoregressive LAM and AdaWorld LAM encode highly temporally redundant actions that may not necessarily compose together to encode the overarching semantic, whereas optical flow only captures pixel level changes between frames and is thus insufficient as well. This is in contrast to the \modelname LAM which places an emphasis on composite actions that must encode the full surgical evolution.

\subsection{Ablation Studies}
First, to further test the assumption that longer horizon modeling in Latent Action Model translates to better ability to capture surgical actions, we train \modelname LAM models for different context lengths $K$ ranging from 2 to 16 while using the same multi-step prediction dynamics model. After the LAM infers $K-1$ actions from $K$ frames, its context is reset to infer further actions. Note that $K = 2$ corresponds to using the AdaWorld LAM. Second, we study how the per frame generation quality varies as the prediction horizon increases for both the single-step and multi-step dynamics model.
% }
\begin{itemize}
    \item \textbf{Impact of context length in latent action model: } 
     As shown in Figure \ref{fig:action_reset}, performance improves significantly with longer horizon modeling in LAM even though the generative modeling ability of the dynamics model is unchanged, which shows that longer horizon action modeling of \modelname LAM is better suited for capturing surgical actions.

    \item \textbf{Generation quality vs prediction horizon:}
    As shown in Figure \ref{fig:per_frame}, both \modelname and AdaWorld show deteriorating performance as the number of generated frames increases. But AdaWorld exhibits faster fall-off in quality from the start, particularly in CMMD, due to its autoregressive nature. We can also see a sudden drop in quality for the 8th generated frame for \modelname, since the model generates 7 frames at a time, with the model resetting its starting context to the 7th frame to generate the additional 8th to 14th frames.
    
\end{itemize}

\section{Conclusion and Future Work}
\label{sec:conclusion}
Our work represents an initial step towards general expertise transfer from surgical demonstrations. Our novel multi-step action learning framework learns better temporal abstractions to generate high fidelity simulations of transferring actions across surgical contexts, which we have validated through extensive experiments. However, it is only applicable to surgical contexts with similar anatomy or view having the same procedure and instrument embodiment. Our work is also limited to unsupervised setting, which, while beneficial for scalability, offers limited insights about potential performance in downstream tasks where ground truth actions are available. Furthermore, the proposed framework utilizes pre-defined latent action trajectory which prevents a truly interactive experience. Future work could explore adapting these models to specific action spaces of specialized surgical tasks.

\clearpage
{
    \small
    \bibliographystyle{ieeenat_fullname}
    \bibliography{main}
}

\clearpage
\setcounter{page}{1}
\maketitlesupplementary

\section{Implementation Details}
\textbf{Latent Action Model}
The encoder takes in frames $f_{0:T}$ as inputs. Each input frame is divided into visual tokens using a patch size of $(16, 16)$ and learnable positional embeddings are applied. The resulting visual tokens are concatenated in time dimension with learnable action prompts $p_{0:T}$ of model dimension. 
After the encoder processes the tokens using causal spatial temporal layers, the first action token is discarded and the rest of the tokens $p_{1:T}$ are used to estimate the posterior distribution of actions: $(\mu_{0:T-1},\,\sigma_{0:T-1})$. We then sample the 32 dimensional action representations $\tilde{a}_{0:T-1}$. For the Multi-Step LAM Model, the sampled actions $\tilde{a}_{0:T-1}$ are concatenated in time dimension with $T-1$ copies of the first frame $f_0$. The resulting tokens are again processed with the causal decoder blocks to get the predicted frames $\hat{f}_{1:T}$. \\

\noindent \textbf{Dynamics Model}
The Stable Video Diffusion model is modified so that in addition to the original conditionings, each frame $f_k$ also conditions on the action $\tilde{a}_{k-1}$ that comes from the LAM encoder. The conditioning on action is done in the form of both cross-attention and scale-shift normalization.
It takes 3.61 seconds to generate 7 frames (i.e. at 1.94 FPS) using single NVIDIA L40 GPU with peak VRAM usage of 15.43 GB. 

\section{Transfer Metrics for Disparate Domains}
Although sampling source and target clips from different videos in the SurgToolLoc dataset might not maintain procedural similarity, we can also evaluate the model's ability to transfer the action trajectories even when the instrument presences and contextual cues are markedly different. 
As shown in Table \ref{tab:transfer_metrics_dd}, \modelname still shows superior performance over the other variants.

\begin{table}[h!]
\centering
\caption{Comparison on Action Transfer for disparate domains}
\label{tab:transfer_metrics_dd}
\resizebox{\linewidth}{!}{ % Adjust 0.9 for smaller/larger table
    \begin{tabular}{p{6.5cm}|cc} % Adjust width if needed
    \toprule
    \textbf{Model} & \textbf{CD-FVD$\downarrow$} & \textbf{ECS$\uparrow$} \\
    \midrule
    AdaWorld~\cite{gao2025adaworld} & 432.45 & 0.416 \\
    Multi-step LAM + Autoregressive Dynamics & 416.278 & 0.418 \\
    \rowcolor{gray!5} Autoregressive LAM + Multi-step Dynamics & 246.316 & 0.471 \\
    \rowcolor{gray!20} \modelname & \textbf{233.455} & \textbf{0.508} \\
    \bottomrule
    \end{tabular}
}
\end{table}

\begin{figure*}
    \centering
    \includegraphics[width=\linewidth]{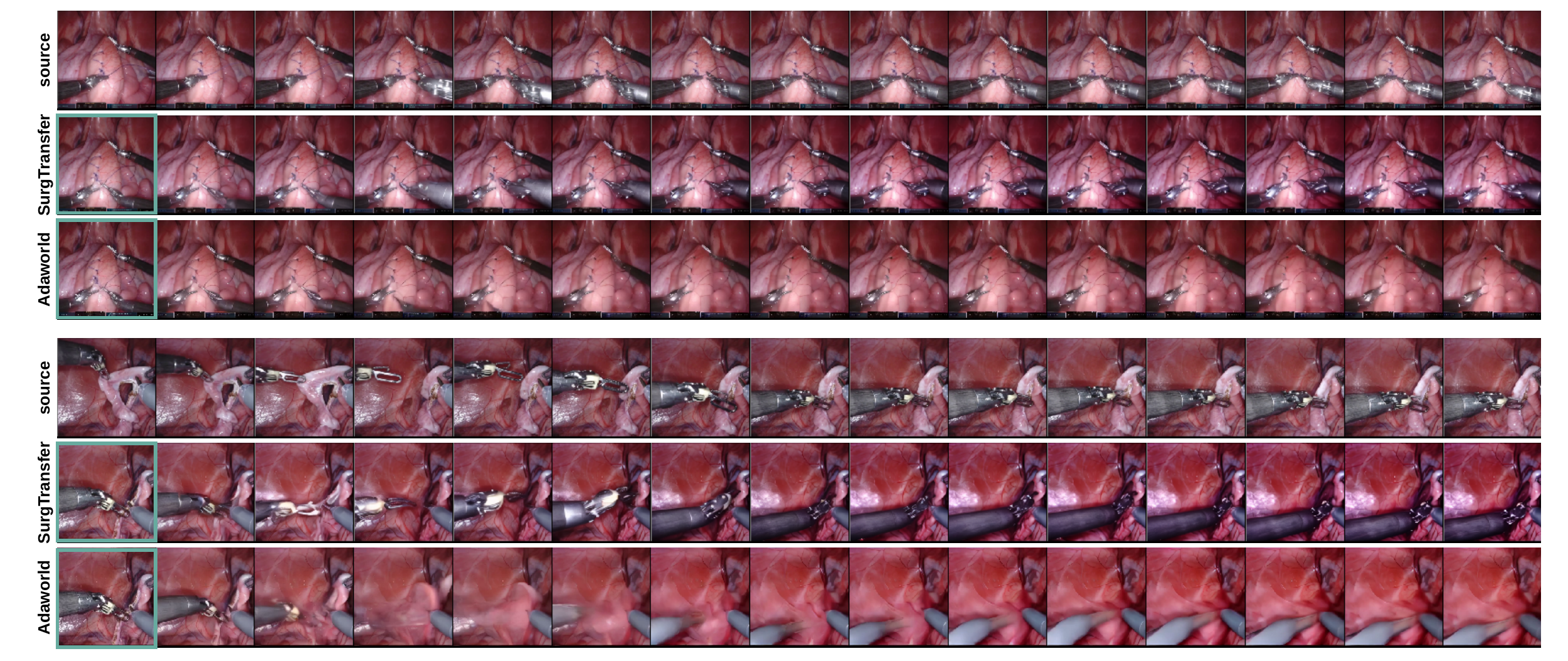}
    \caption{Action transfer (source and target from different parts of same video): source represents the source clip from which the action is inferred, the first frame represents the context, rest of the frames are generated using the context and transferred actions. In the first example, for \modelname, we can see the needle driver on the bottom right holding the thread as in the source video. In the second example \modelname replicates the grasping and the subsequent release of a tissue by bipolar forceps as opposed to AdaWorld.}
    \label{fig:more_same_transfer}
\end{figure*}

\begin{figure*}
    \centering
    \includegraphics[width=\linewidth]{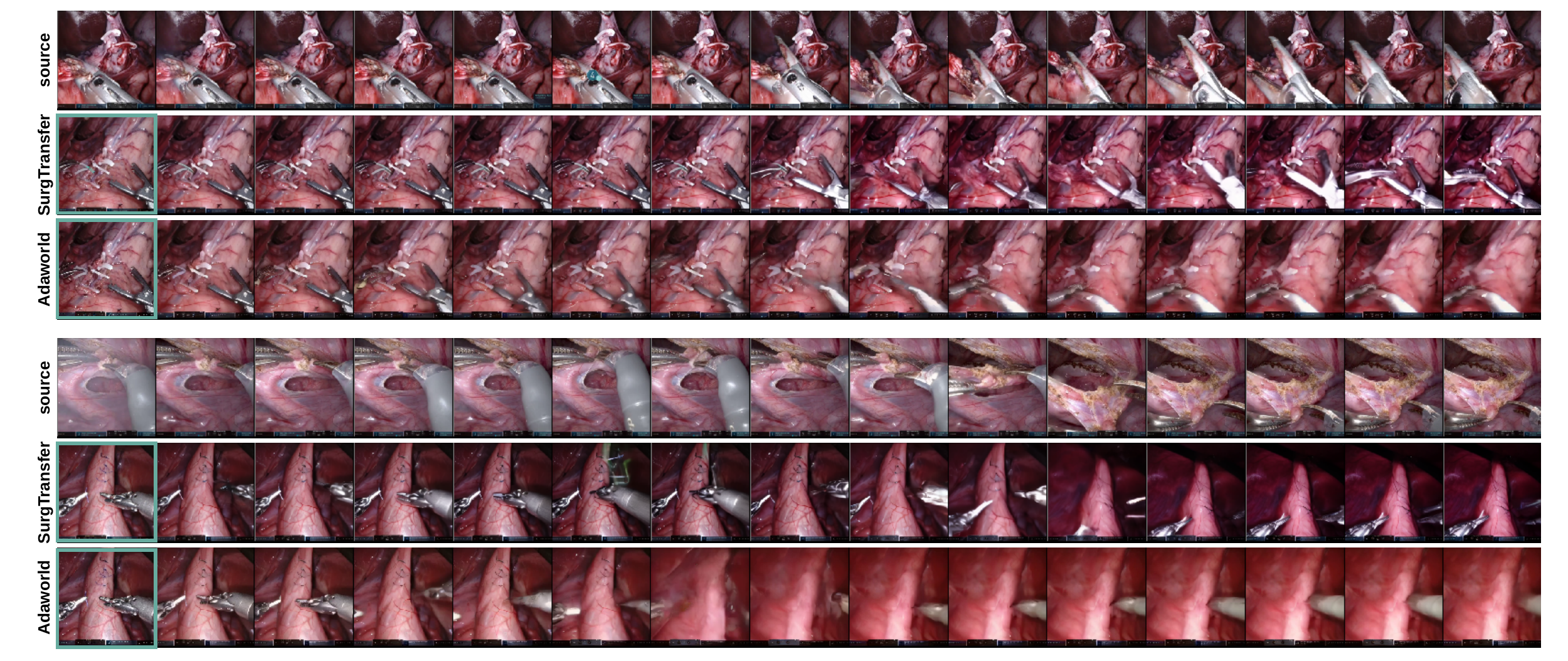}
    \caption{Action transfer (source and target taken from different videos): source represents the source clip from which the action is inferred, the first frame represents the context, rest of the frames are generated using the context and transferred actions. In the first example, both models show increase in brightness for the visible instrument in later frames as in the source video, showing that the models have learned to transfer visual intensity as well. In the second example, the motion of monopolar curved scissors in the source video is transferred to the needle driver present in the target context. Here \modelname can more effectively replicate the sudden movements of the instrument than AdaWorld.}
    \label{fig:more_diff_transfer}
\end{figure*}

\section{Additional Qualititative Results}
We present additional samples for \modelname and AdaWorld of action transfers across same and different videos on the SurgToolLoc dataset in figures \ref{fig:more_same_transfer} and \ref{fig:more_diff_transfer} respectively. 

\section{Clinical Evaluation Details}
We provide the full questionnaire used for clinical evaluation of the action transfer in this section.

\subsection{Instructions:}

\textit{You will be shown a real surgical video (source) and a surgical video generated by an AI model (target) by following the source actions in a new target context.}

\textit{Please assess whether the AI-generated target video preserves surgical reality, causality and intent while performing the transformation.}

\subsection{Visual Perceptual Plausibility}

\textit{Does this transformed video still look like real surgery?}

\begin{enumerate}
	\item The target video is visually clear, stable and free from distracting artifacts 
	\item Lighting, color balance and camera behavior are consistent with a real operative setting
	\item Tissue textures and surfaces in the target video appear realistic.
	\item Temporal continuity is preserved without flicker, jitter or visual discontinuity.
\end{enumerate}

\subsection{Instrument Operation Plausibility}

\textit{Do instruments still behave like surgical instruments after transfer?}

\begin{enumerate}
	\item Instruments in the target video are anatomically and procedurally appropriate
	\item Instrument trajectories and movements remain physically plausible
	\item No instrument behavior appears newly introduced or corrupted by the transformation
	\item Instrument co-ordination reflects realistic surgical technique
\end{enumerate}

\subsection{Environment feedback plausibility}

\textit{Does tissue respond correctly to action in the transformed video?}

\begin{enumerate}
	\item Tissue deformation in the target video corresponds appropriately to instrument interaction
	\item Bleeding, fluid dynamics or coagulation responses (if present) are realistic
	\item Anatomical relationships remain consistent with known surgical anatomy
	\item No tissue response violates physical or medical common sense
\end{enumerate}

\subsection{Surgical intent plausibility}

\textit{Is the surgical goal preserved across the transformation?}

\begin{enumerate}
	\item The overall surgical intent in the target video matches that of the source video
	\item The procedural stage implied by the target video is appropriate
	\item The sequence of actions reflects coherent surgical reasoning rather than visual mimicry
	\item No new or conflicting surgical intent has been introduced by the AI transformation
\end{enumerate}

\subsection{Source-target fidelity}

\textit{Is the surgical goal preserved across the transformation?}

\begin{enumerate}
    \item Key anatomical structures are preserved between source and target videos.
    \item Instrument identity and functional role are preserved across the transformation.
    \item The temporal order of surgical actions is preserved.
    \item No clinically meaningful information appears lost or distorted in the target video.
\end{enumerate}

\section{Algorithm}
We provide the detailed algorithms for training the models and performing action transfer in algorithms~\ref{alg:lam_training}, \ref{alg:dynamics_training}, and \ref{alg:action_transfer}.

% ===================== Stage 1 =====================

\begin{algorithm}[t]
\caption{Multi-step Latent Action Learning}
\label{alg:lam_training}
\begin{algorithmic}[1]
\Require Unlabeled surgical videos $\mathcal{V}$; prediction window $T$; LAM encoder $\mathcal{E}_{\phi}$; LAM decoder $\mathcal{D}_{\psi}$; KL weight $\beta$

\For{each training iteration}
    \State Sample a clip $f_{0:T} = \{f_0, f_1, \ldots, f_T\}$ from $\mathcal{V}$
    \For{$t = 1$ to $T$}
        \State Infer multi-step latent actions
        \[
            \tilde{a}_{0:t-1} \sim q_{\phi}\!\left(a_{0:t-1}\mid f_0, f_{1:t}\right)
            =
            \mathcal{E}_{\phi}(f_0, f_{1:t})
        \]
        \State Reconstruct the future using only the initial context and the inferred actions
        \[
            \hat{f}_{1:t}
            =
            \mathcal{D}_{\psi}(f_0, \tilde{a}_{0:t-1})
        \]
        \State Compute the multi-step LAM objective
        \[
        \begin{aligned}
        \mathcal{L}_{\mathrm{LAM}}
        &=
        \left\| f_{1:t} - \hat{f}_{1:t} \right\|_2^2 \\
        &\quad+
        \beta
        D_{\mathrm{KL}}
        \left(
            q_{\phi}(\tilde{a}_{0:t-1}\mid f_0, f_{1:t})
            \,\|\, \mathcal{N}(0, I)
        \right)
        \end{aligned}
        \]
    \EndFor
    \State Update $\phi,\psi$ by minimizing $\mathcal{L}_\mathrm{LAM}$
\EndFor

\end{algorithmic}
\end{algorithm}

% ===================== Stage 2 =====================

\begin{algorithm}[t]
\caption{Multi-step Action-conditioned Dynamics Learning}
\label{alg:dynamics_training}
\begin{algorithmic}[1]

\Require Unlabeled surgical videos $\mathcal{V}$; LAM encoder $\mathcal{E}_{\phi}$; diffusion dynamics model $\mathcal{G}_{\theta}$

\For{each training iteration}
    \State Sample a clip $f_{0:T}$ from $\mathcal{V}$

    \State Infer the composite latent action trajectory
    \[
        \tilde{a}_{0:T-1}
        =
        \mathcal{E}_{\phi}(f_0, f_{1:T})
    \]

    \State Let $v = f_{1:T}$ be the target future video

    \State Sample diffusion noise $\epsilon$ and noise level $\tau$, and construct $v_{\mathrm{noised}}$

    \State Predict the denoised video with the dynamics model
    \[
        \hat{v}
        =
        \mathcal{G}_{\theta}
        \left(
            v_{\mathrm{noised}}, f_0, \tilde{a}_{0:T-1}, \tau
        \right)
    \]

    \State Update $\theta$ by minimizing
    \[
        \mathcal{L}_{\mathrm{dyn}}
        =
        \mathbb{E}_{v,\epsilon,\tau}
        \left[
            \left\|
                v - \hat{v}
            \right\|_2^2
        \right]
    \]

\EndFor

\end{algorithmic}
\end{algorithm}

% ===================== Stage 3 =====================

\begin{algorithm}[t]
\caption{Cross-context Surgical Action Transfer}
\label{alg:action_transfer}
\begin{algorithmic}[1]

\Require Source demonstration $f^{s}_{0:H}$; target context frame $f^{t}_{0}$; trained LAM encoder $\mathcal{E}_{\phi}$; diffusion dynamics model $\mathcal{G}_{\theta}$ with $T$ being the number of generated frames per single pass of the diffusion model.

\Ensure Transferred target video $\hat{f}^{t}_{1:H}$

\State Initialize context frame ${f}^{c} \leftarrow f^{t}_{0}$

\For{each prediction block $b = 0,1,\ldots,\lceil H/T\rceil-1$}

    \State Extract source block $f^{s}_{bT:(b+1)T}$

    \State Infer source latent actions
    \[
        \tilde{a}^{s}_{b}
        =
        \mathcal{E}_{\phi}
        \left(
            f^{s}_{bT},
            f^{s}_{bT+1:(b+1)T}
        \right)
    \]

    \State Generate the corresponding target future block
    \[
        \hat{f}^{t}_{bT+1:(b+1)T}
        =
        \mathrm{Sample}
        \left(
            \mathcal{G}_{\theta},
            {f}^{c},
            \tilde{a}^{s}_{b}
        \right)
    \]

    \State Set the last generated frame as the context for the next block
    \[
        {f}^{c}
        \leftarrow
        \hat{f}^{t}_{(b+1)T}
    \]

\EndFor

\State \Return transferred target video $\hat{f}^{t}_{1:H}$

\end{algorithmic}
\end{algorithm}

\section{Latent Action Information Visualization}
\begin{figure*}
    \centering
    \includegraphics[width=\linewidth]{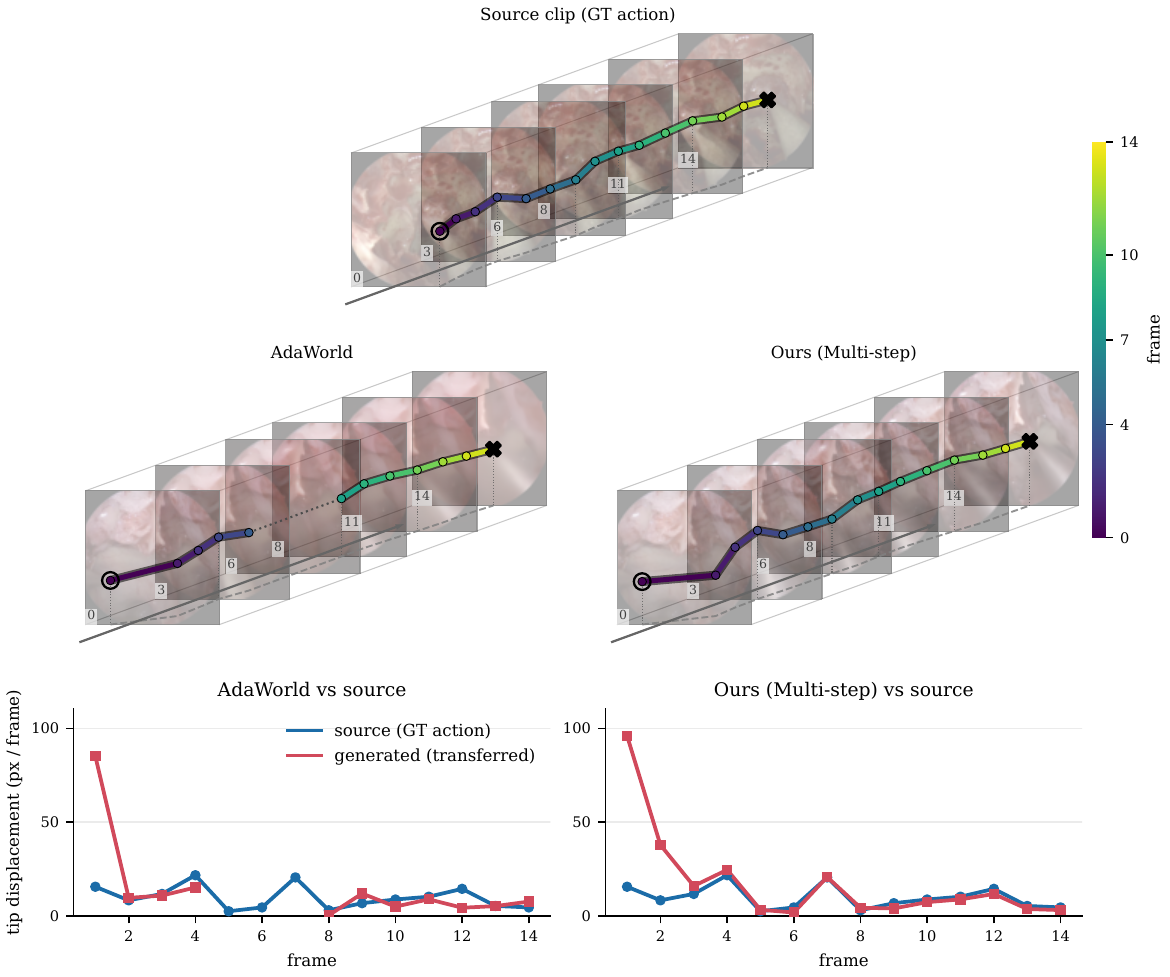}
    \caption{Here, \modelname tightly tracks the relative tip displacements showing high adherence to the conditioning action signal.}
    \label{fig:latent_good}
\end{figure*}

\begin{figure*}
    \centering
    \includegraphics[width=\linewidth]{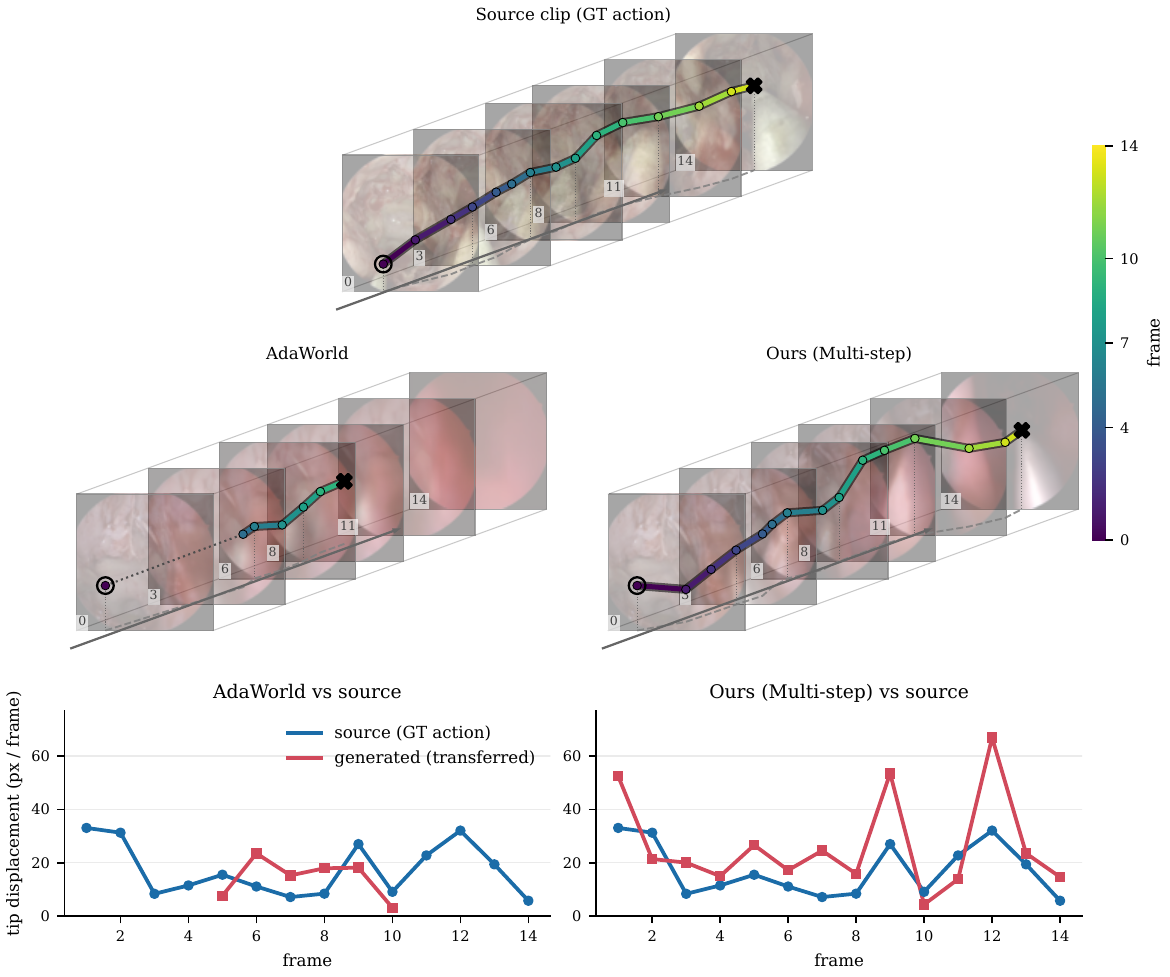}
    \caption{Although \modelname has high error in tracking tip displacement, we can see that the error is due to over-amplification of the movements in the source video suggesting that the model can be prone to overweighing the conditioning action signal.}
    \label{fig:latent_more}
\end{figure*}

We provide further examples of instrument trajectory visualization in figures~\ref{fig:latent_good}, and ~\ref{fig:latent_more}.

\end{document}